\documentclass{bioinfo}

\usepackage{multirow}
\usepackage{subfigure}

\copyrightyear{2022} \pubyear{2022}

\access{Advance Access Publication Date: Day Month Year}
\appnotes{Manuscript Category}

\begin{document}
\firstpage{1}

\subtitle{Data and text mining}

\title[Benchmark for Automatic Medical Consultation System]{A Benchmark for Automatic Medical Consultation System: Frameworks, Tasks and Datasets}
\author[Wei Chen \emph{et~al}.]{Wei Chen\,$^{\text{\sfb 1}}$, Zhiwei Li\,$^{\text{\sfb 1}}$, Hongyi Fang\,$^{\text{\sfb 1}}$, Qianyuan Yao\,$^{\text{\sfb 1}}$, Cheng Zhong\,$^{\text{\sfb 1}}$, Jianye Hao\,$^{\text{\sfb 3}}$, Qi Zhang\,$^{\text{\sfb 4}}$, Xuanjing Huang\,$^{\text{\sfb 4}}$, Jiajie Peng\,$^{\text{\sfb 2,5,}*}$ and Zhongyu Wei\,$^{\text{\sfb 1,5,}*}$}
\address{$^{\text{\sf 1}}$School of Data Science, Fudan University, Shanghai, 200433, China  and \\
$^{\text{\sf 2}}$School of Computer Science, Northwestern Polytechnical University, Xi'an, 710000, China  and \\
$^{\text{\sf 3}}$College of Intelligence and Computing, Tianjin University, Tianjin, 300072, China and \\ 
$^{\text{\sf 4}}$School of Computer Science, Fudan University, Shanghai, 200433, China and \\ 
$^{\text{\sf 5}}$Research Institute of automatic and Complex Systems, Fudan University, Shanghai, 200433, China.}

\corresp{$^\ast$To whom correspondence should be addressed.}

\history{Received on XXXXX; revised on XXXXX; accepted on XXXXX}

\editor{Associate Editor: XXXXXXX}

\abstract{\textbf{Motivation:} In recent years, interest has arisen in using machine learning to improve the efficiency of automatic medical consultation and enhance patient experience. In this article, we propose two frameworks to support automatic medical consultation, namely doctor-patient dialogue understanding and task-oriented interaction. We create a new large medical dialogue dataset with multi-level fine-grained annotations and establish five independent tasks, including \emph{named entity recognition}, \emph{dialogue act classification}, \emph{symptom label inference}, \emph{medical report generation} and \emph{diagnosis-oriented dialogue policy}.\\
\textbf{Results:} We report a set of benchmark results for each task, which shows the usability of the dataset and sets a baseline for future studies.\\
\textbf{Availability:} Both code and data is available from \href{https://github.com/lemuria-wchen/imcs21}{https://github.com/lemuria-wchen/imcs21}. \\
\textbf{Contact:} \href{chenwei18@fudan.edu.cn}{chenwei18@fudan.edu.cn}\\
\textbf{Supplementary information:} Supplementary data are available at \emph{Bioinformatics}
online.}

\maketitle

\section{Introduction}

Online medical consultation has shown great potential in improving the quality of healthcare services while reducing cost \citep{singh2018online}, especially in the era of raging epidemics such as \emph{Coronavirus} \citep{singhal2020review}. This fact has accelerated the emergence of online medical communities like \emph{SteadyMD} (https://www.steadymd.com) and \emph{Haodafu} (https://www.haodf.com/). These platforms provide a medium for doctors and patients to communicate with each other remotely, which is called telemedicine \citep{wootton2001telemedicine}.

\begin{figure}
\centering{
\includegraphics[width=0.8\columnwidth]{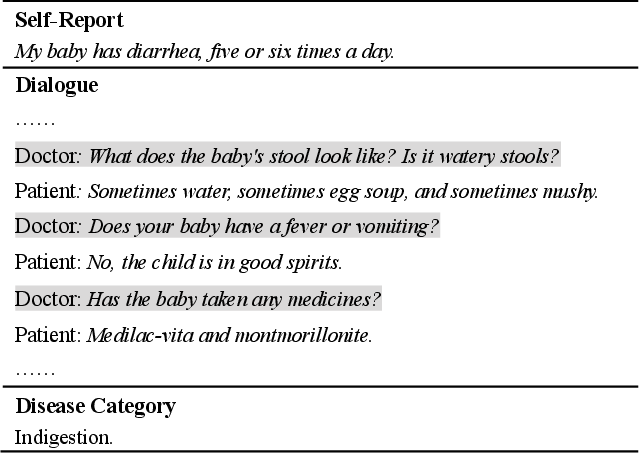}}
\caption{An example of medical consultation record, where the text is translated from Chinese.}
\label{fig:mcr_sample}
\end{figure}

Typically, in telemedicine, the patient first provides a brief summary of their physical condition, i.e., self-report, then the doctor communicates with the patient to learn more about the patient's health condition. After sufficient inquiry, the doctor may make a diagnosis and provide further medical advice. The electronic record of this process is called Medical Consultation Record (MCR). Figure \ref{fig:mcr_sample} demonstrates an example of MCR, which consists of patient's self-report, plain text of dialogue and corresponding disease category. 

Recently, researchers have paid close attention to develop automatic approaches to facilitate online consultation service. Research topics include medical named entity recognition \citep{zhou2021end}, drug recommendation \citep{zheng2021drug}, automatic text-based diagnosis \citep{chen2020towards}, health question answering \citep{he-etal-2020-infusing}, medical report generation \citep{joshi-etal-2020-dr} and diagnostic policy \citep{wei2018task,10.1093/bioinformatics/btac744}. Although progresses have been made to support automatic medical consultation from different perspectives, there is still a large gap between existing work and real-world application. We summarize this gap to two major limitations: 1) Lack of design of frameworks and tasks for automatic medical consultation; 2) Lack of benchmark datasets to support the development of research and application. 



In this article, we make the first step to build a framework for automatic medical consultation and propose several tasks to cover the entire procedure. Two modes of frameworks are proposed to support both static and dynamic scenarios, namely, \emph{dialogue understanding} and \emph{task-oriented interaction}. The understanding framework aims to extract structured information from the dialogue context and generate useful labels to describe the dialogue state, which can include the patient's health status, patient's intention, etc. The interaction framework is designed to learn the dialogue policy, that is, to select the next action based on the current dialogue state, such as asking the patient whether he or she has a certain symptom. We create a corpus called IMCS-21 with multi-level fine-grained annotations to support the research and application development of five tasks under the two modes. We develop widely used neural-based models for each task and report a set of benchmark results, which shows the usability of the corpus and sets a baseline for future studies. We conduct a comprehensive analysis of our corpus and tasks to show great future opportunities. 








The \emph{main contributions} of this article can be summarized as follows: 1) We propose a design of frameworks and tasks for automatic medical consultation and introduce IMCS-21, a large-scale annotated medical dialogue corpus, whose superiority makes it potentially a great benchmark for medical dialogue modeling; 2) We created neural-based models for each task and report a set of benchmark results. We will continue to track the progress of these tasks.





\section{Related Work}
\label{section:related}

To build an automatic medical consultation system, learning from a large amount of actual doctor-patient conversations and directly imitate human behavior may be the best strategy. There are already a few medical dialogue corpus introduced by previous studies. These corpus can be roughly divided into two categories. 

One such category is original medical conversations corpus between patients and doctors with no annotations. MedDialog \citep{zeng2020meddialog} is a large scale medical dialogue dataset that contains a Chinese dataset with 3.4 million conversations covering 172 specialties of diseases and an English dataset with 0.26 million conversations covering 96 specialties of diseases. KaMed \citep{li2021semi} is a knowledge aware medical dialogue dataset that contains over 60,000 medical dialogue sessions and is equipped with external medical knowledge from Chinese medical knowledge platform. The tasks built on these corpus are usually response generation in dialogue systems, on which researchers can build automated medical chatbots. However, the responses generated by such end-to-end chatbots lack interpretability and controllability, which has strong limitations in healthcare applications. 

Another category is the annotated doctor-patient medical dialogue corpus. The annotated content of these corpus is related to the task they focus on and the researchers establish a series of medical dialogue modeling tasks including natural language understanding (NLU), natural language generation (NLG) and dialogue policy (DP). MSL \citep{shi2020understanding} is a dataset for slot filling task which aims to transform a natural language medical query in which colloquial expressions exist into the formal representation with discrete logical forms to perform correct query. CMDD \citep{lin2019enhancing}, SAT \citep{du-etal-2019-extracting} and MIE \citep{zhang2020mie} are datasets for medical information extraction task, which is extract mentioned entities and their corresponding status. MZ \citep{wei2018task}, DX \citep{xu2019end}, RD/SD \citep{zhong2022hierarchical} are datasets that contain structured symptom features to learn the dialogue policy for symptom based automatic diagnosis. Chunyu \citep{lin2021graph} is a dataset for end-to-end diagnosis-oriented response generation task. MedDG \citep{liu2022meddg} contains more than 17K conversations with annotated entities, and two medical dialogue tasks are established. One is the next sentence entity prediction and the other is the dialogue response generation. 

One challenge of existing datasets is the medical label insufficiency. The majority of datasets only provide one specific medical labels, e.g., medical entities. These labels are too coarse to accurately describe the patient's state and intent. Another challenge is the small scale of existing annotated datasets, typically on the order of hundreds of dialogues. In the supplementary material, we present the comparative details between IMCS-21 and existing medical (dialogue) datasets.

\section{Automatic Medical Consultation: Frameworks and Tasks}
\label{section:frame}

\begin{figure}
\centering{
\includegraphics[width=0.8\columnwidth]{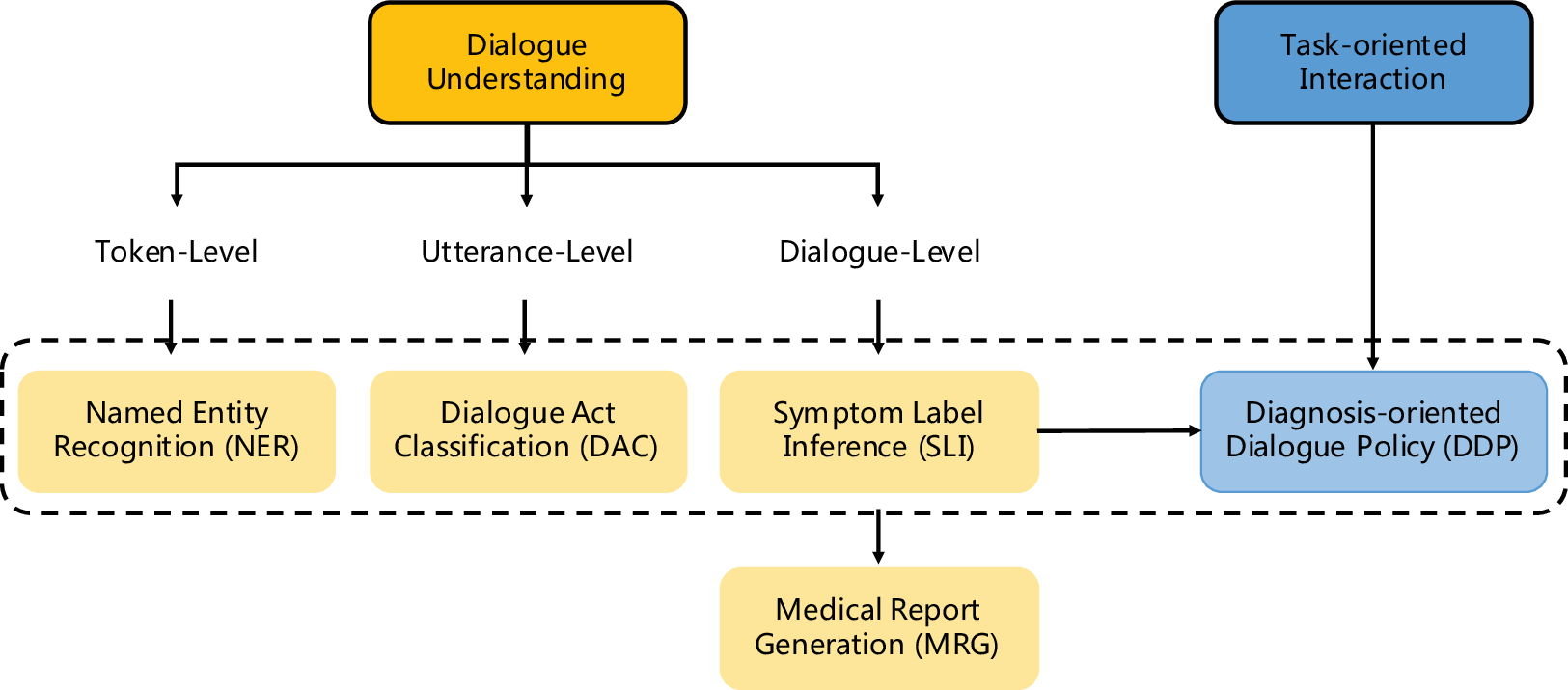}}
\caption{Framework and task design for automatic medical consultation.}
\label{fig:frame}
\end{figure}

We present our design of frameworks and tasks for automatic medical consultation system in Figure \ref{fig:frame}. 


\subsection{Dialogue Understanding Framework}

The understanding framework includes 4 tasks: Named Entity Recognition (NER), Dialogue Act Classification (DAC), Symptom Label Inference (SLI) and Medical Report Generation (MRG). 

\paragraph{Named Entity Recognition} ~ {Medical NER task aims to recognize pre-defined medical named entities from medical texts \citep{zhou2021end}. Medical related entities are widely present in actual doctor-patient conversations, and NER is a basic task for extracting medical semantics.}

\paragraph{Dialogue Act Classification} ~ {DAC is the task of classifying an utterance with respect to the function it serves in a dialogue, i.e. the act the speaker is performing \citep{liu2017using}. In medical dialogues, the identification of dialogue act is an important aspect in analyzing the doctor's and patient's intent and what they are trying to convey. }

\paragraph{Symptom Label Inference} ~ {Symptoms are the main topics discussed in medical dialogues and an important basis for doctors to make a diagnosis \citep{lin2019enhancing}. The goal of SLI task is to identify mentioned symptoms from the dialogue, align them to standardized names, and determine whether a patient suffers from these symptoms. SLI task generates a clearer structured symptom features about the patient.}


\paragraph{Medical Report Generation} ~ {Medical report captures and summarizes the important parts of the medical conversation needed for clinical decision making and subsequent follow ups \citep{joshi-etal-2020-dr}. As a way to record and convey medical information, MRG task addresses a practical need and plays an important role in medical practice. }

\subsection{Task-oriented Interaction framework}

The interaction framework controls the process of man-machine dialogue, which is to determine the future dialogue strategy based on historical information. For interaction framework, we introduce Diagnosis-oriented Dialogue Policy (DDP) task which follows the setting of task-oriented dialogue system \citep{wei2018task}. 


\paragraph{Diagnosis-oriented Dialogue Policy} ~ {The DDP task aims to learn the optimal policy for symptom-based automatic disease diagnosis. The policy is expected to efficiently find potential symptoms of patients and make a correct diagnosis, through several turns of interaction. It is worth noting that the training data required for DDP is exactly the structured symptom features the SLI task needs to predict.}

\section{Medical Dialogue Corpus: IMCS-21}
\label{section:corpus}

In this section, we present our collection and analysis of the annotated dataset. The raw data comes from Muzhi (http://muzhi.baidu.com), a Chinese online health community that provides professional medical consulting service for patients. We collect extensive medical consultation records for 10 pediatric diseases. After removing some incomplete samples and samples with too short dialogues, we annotate the filtered samples to form our our medical dialogue corpus, which we call IMCS-21.

\subsection{Annotation Scheme}

The annotation scheme is designed by medical experts with consideration of our task design as well as actual scenarios of online consultation. 

Specifically, we collect multi-level annotations for each medical consultation record, including token level, utterance level, and dialogue level. At token level, the annotator is asked to find medical named entities; at utterance level, the intention of each utterance is annotated; at dialogue level, the symptom features are collected, and the medical report is manually written. 

\paragraph{Medical Named Entity} ~ {We define 5 main categories of entities for annotation after discussing with domain experts, i.e., \emph{symptom (SX)} , \emph{drug name (DN)}, \emph{drug category (DC)}, \emph{examination (EX)} and \emph{operation (OP)}. These categories of entities we believe are important for understanding the doctor-patient dialogue. Among them, \emph{drug name} represents a specific drug name while \emph{drug category} represents a class of drugs with a certain efficacy. For example, ``aspirin" belongs to \emph{DN} while ``anti-inflammatory drug" belongs to \emph{DC}. Besides, \emph{OP} represents related medical operations, such as ``infusion", ``atomization", etc.


Inside–outside–beginning (BIO) \citep{ramshaw1999text} tagging scheme is employed, where ``B'' and ``I'' determine the boundary of an entity. For each Chinese character, ``B'' stands for the beginning of the entity, ``I'' means inside, and ``O'' means other. This results in 11 possible tags for tokens. We assign an initial label to each token using a rule-based algorithm \citep{aho1975efficient} to prompt the annotation process.}

\paragraph{Dialogue Act} ~ {The categories of dialogue acts are determined according to the specific content of the utterance. It can be broadly divided into two big categories: \emph{request (R)} and \emph{inform (I)}, one means ``\emph{ask for information}", and another means ``\emph{tell the information}". We further categorize the content of information conveyed as: \emph{basic information (BI)}, \emph{symptom (SX)}, \emph{etiology (ETIOL)}, \emph{existing exam and treatment (EET)}, \emph{medical advice (MA)}, \emph{drug recommendation (DR)}, \emph{precautions (PRCTN)}, \emph{diagnose (DIAG)}. There are both request and inform versions for all categories except \emph{DIAG}. Utterances that does not fall into the above categories will be labeled as \emph{other (OTR)}. 

This results in a total of 16 possible categories of dialogue acts in our scheme. Each utterance in a dialogue is tagged with one of these categories. In this article, we use abbreviations to denote specific dialogue acts. For example, \emph{R-SX} is abbreviate for \emph{request-symptom}, which represents the intent of asking someone for relevant symptoms. To be more intuitive, we demonstrate several examples for each entity category and dialogue act category in the supplementary material.}



\paragraph{Symptom Label} ~ {In order to clarify the relationship between the symptoms appearing in the dialogue and the patient, each symptom entity is additionally tagged with a label: \emph{Positive (POS)}, \emph{Negative (NEG)} or \emph{Not Sure (NS)}, to indicate whether the patient has the symptom. The symptom label determines the relationship between the symptom and the patient. The annotator can infer the symptom label by observing the utterance where the symptom entity is located and its context. Besides, all identified symptoms are normalized by linking them to the most relevant one on SNOMED-CT2 (https://www.snomed.org/snomed-ct), which can unify different expressions of the same symptom into one standard name. Symptoms mentioned in self-report are also identified and normalized.}


\paragraph{Medical Report} ~ {Annotators are also required to write a report in specified format to summarize the medical consultation case. It contains six parts: 1) \emph{chief complaint}: patient's main symptoms or signs; 2) \emph{present disease}: description of main symptoms; 3) \emph{auxiliary examination}: the patient's existing examinations, examination results, records, etc; 4) \emph{past medical history}: previous health conditions and illnesses; 5) \emph{diagnosis}: diagnosis of disease; 6) \emph{suggestions}: doctor's suggestions of inspection recommendations, drug treatment and precautions. Annotators are required to fill in these parts and leave it blank if the part is not mentioned in the dialogue.}

\subsection{Inter-Annotator Agreement}

For the annotation of medical dialogues, we develop a web-based tool which can be utilized for general-purpose multi-turn dialogue labeling. We recruit undergraduates and postgraduates in medical school to annotate our corpus. All annotators are people who are willing to participate and over the age of 18. 

Two annotations per dialogue are gathered, and inconsistent parts are further finalized by a third annotator. We use Cohen's kappa coefficient \citep{banerjee1999beyond} to estimate the inter-annotator agreement. For the annotations of medical named entities and dialogue acts, the kappa coefficients are 83.11\% and 76.41\% respectively; For the annotations of symptom labels, the kappa coefficients is 92.71\%; For medical reports, both reports are remained for reference. These results show that the consistency between the annotators is satisfactory. 

\subsection{Corpus Analysis}

\begin{table}
\centering
\small
\caption{Statistics of IMCS-21.}
\label{tab:stats}
\begin{tabular}{lr}  \toprule
\textbf{Statistics} & \textbf{Avg.} \\\midrule
\# of utterances per dialogue  & 40    \\
\# of characters per utterance      & 523   \\
\# of characters per self-report    & 57    \\
\# of entities per dialogue (annotated)    & 26    \\
\# of characters per medical report (annotated) & 88    \\  \botrule
\end{tabular}
\end{table} 

\paragraph{Corpus Statistics} ~ {IMCS-21 contains a total of 4,116 annotated samples with 164,731 utterances, which covers 10 pediatric diseases: \emph{bronchitis}, \emph{fever}, \emph{diarrhea}, \emph{upper respiratory infection}, \emph{dyspepsia}, \emph{cold}, \emph{cough}, \emph{jaundice}, \emph{constipation} and \emph{bronchopneumonia}. Each dialogue contains an average of 40 utterances, 523 Chinese characters (580 characters if including self-report) and 26 entities (see in Table \ref{tab:stats}).}

\begin{figure*}
\centering
\small
\subfigure[\# of entity categories]{
\label{fig:et_dist}
\includegraphics[width=0.6\columnwidth]{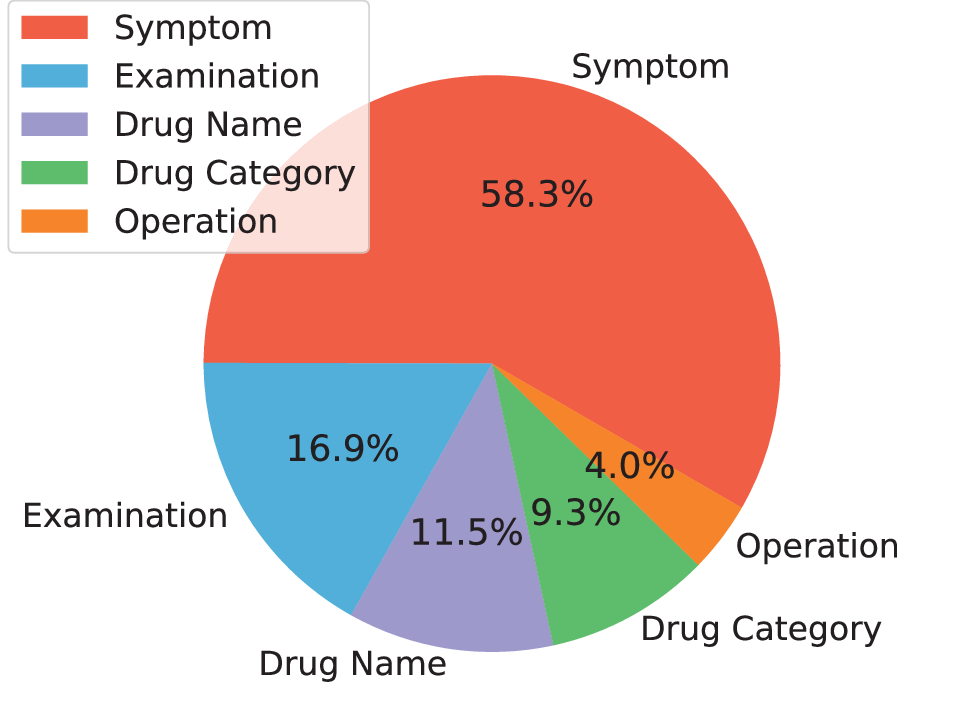}}
\subfigure[\# dialogue acts]{
\label{fig:da_dist}
\includegraphics[width=0.6\columnwidth]{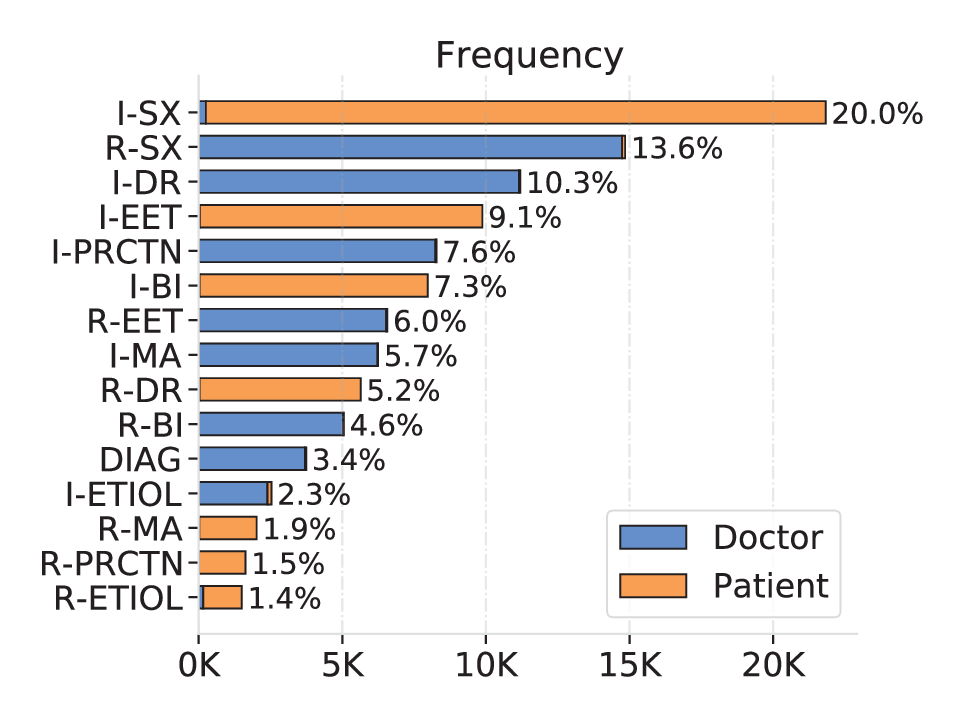}}
\subfigure[position of dialogue acts]{
\label{fig:da_pos}
\includegraphics[width=0.6\columnwidth]{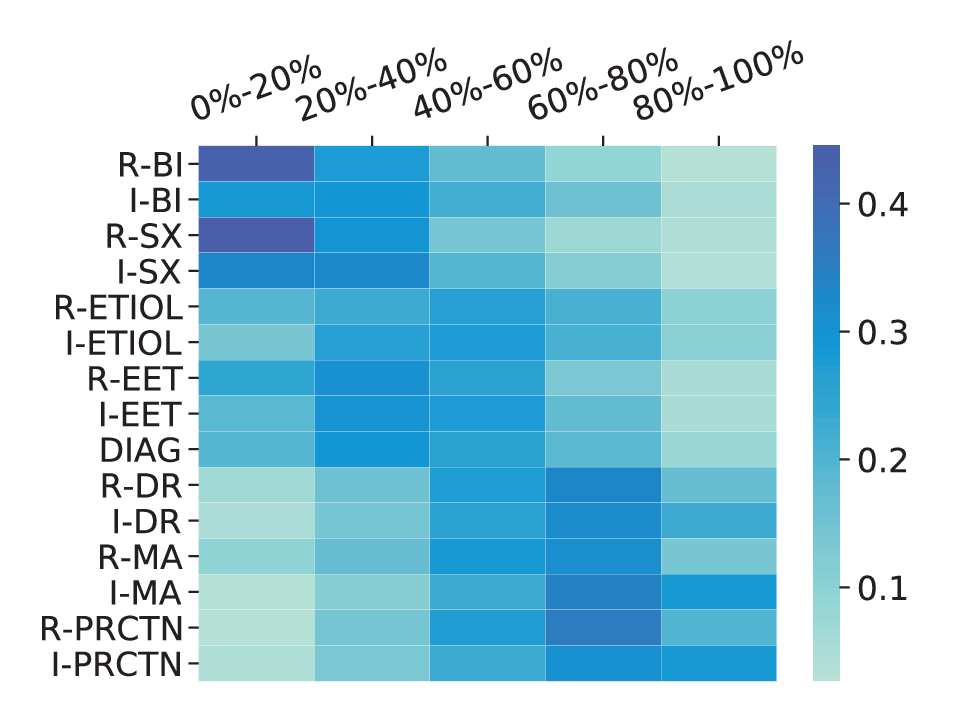}}
\caption{Pie chart for number of entity categories, and bar chart and heat-map for the number and locations of dialogue acts respectively. Note that we exclude category \emph{other} in the statistics of dialogue acts.}
\label{fig:dist}
\end{figure*}

\paragraph{Dialogue Content Analysis} ~ {The distribution of number of entity categories and dialogue act categories are shown in Figure \ref{fig:et_dist} and \ref{fig:da_dist}. Briefly, \emph{symptom} entities appear the most in conversations, about 58.3\%. Similarly, the two categories with the highest proportion of dialogue acts are \emph{I-SX} and \emph{R-SX}. This indicates that doctor-patient conversations mainly discuss the patient's symptoms. Examinations, drugs, advice and precautions are also common topics, this suggests that patients try to find medical solutions in consultations. 

It is worth noting that for any specific category of dialogue act, it is either almost from doctors or patients (Figure \ref{fig:da_dist}). However, there are some exceptions. For example, the category \emph{I-SX} means telling the other about the symptoms, which intuitively will only come from the patient who tells the doctor his symptom. But sometimes the doctor may remind the patient what symptoms they actually have, based on the their previous vague description. For example, the utterance ``\emph{your body temperature is relatively high, it is febrile}'' from the doctor will be labeled as \emph{I-SX}.}

\paragraph{Dialogue Structure Analysis} ~ {Figure \ref{fig:da_pos} present the positional distribution characteristics of dialogue act categories}. We divide utterances in a dialogue into five parts according to their locations. For example, 0-20\% means the sentences appeared in the first fifth of the conversation. From the Figure, we conclude that with the in-depth of medical consultation, the focus gradually shifts from symptoms to drugs, treatments and precautions. Clearly, there are certain regularities in the structure of medical dialogues for the purpose of diagnosis.

\paragraph{Symptoms Analysis} ~ {We present the statistics of symptoms and symptom labels in Table \ref{tab:symptom}. Each self-report and dialogue contains 1.7 and 6.6 (unique) symptom entities on average. In the dialogue, the number of non-positive symptom entities account for nearly 40\%, which means that a large proportion of the symptoms in the conversation may not be related to the patient.}

\begin{table}
\centering
\small
\caption{Statistics of symptoms and symptom labels.}
\label{tab:symptom}
\begin{tabular}{cccc} \toprule
                & \textbf{Self-Report} & \textbf{Dialogue (POS / NEG / NS)} & \textbf{Total} \\ \midrule
\# of Symptoms & 1.7        & 4.0 / 1.6 / 1.0        & 8.3 \\ \botrule
\end{tabular}
\end{table}

\paragraph{Medical Reports Analysis}\label{section:mra} ~ {In the annotation of medical reports, without being provided with true disease labels, the annotators are required to populate the \emph{diagnosis} part with the patient's disease they infer from the conversation. Therefore, the accuracy of the content of this part can roughly assess how well the annotator understands the dialogue. By regex matching, we find that in 84.7\% of the reports, the content of the \emph{diagnosis} part contains the text of the actual disease or the key concepts, which ensures the quality of medical reports acceptably. 

It is worth noting that some diseases are hard to distinguish from others, or are themselves a symptom of other diseases. In this case, annotators are easily confused. For example, when the real disease is \emph{Cold}, only 65.6\% of the reports contain the key concepts of cold in the \emph{diagnosis} part. When the disease is \emph{Jaundice}, the proportion is as high as 98.1\%. 

Besides, the \emph{Present disease} and \emph{Suggestions} part has about 30 and 20 words on average respectively, which occupy the main content of medical reports, while a considerable percentage (about 60\%) of \emph{past medical history} is empty, because this part is less involved in the dialogue.}

\section{IMCS-21 as a New Benchmark}
\label{section:benchmark}

As introduced in Section \ref{section:frame}, we break down the medical consultation modelling into two modes of frameworks, comprising a total of five tasks. To show the potential usefulness of IMCS-21, we establish a standard split for IMCS-21 at the dialogue level, and report a benchmark result for each of the task: NER, DAC, SLI, MRG and DDP. The split is consistent across all tasks, consisting of a training set with 2,472 dialogues, a develop set with 833 dialogues and a test set with 811 dialogues. 

\begin{table}
\centering
\small
\caption{List of notations for task formalization. The \emph{dimension} column denotes the size of the set represented by these notations in our task settings. There are three elements in $\mathcal{L}$, namely \emph{POS, NEG and NS}.}
\label{tab:notation}
\begin{tabular}{llr} \toprule
\textbf{Sign}         & \textbf{Description}                    & \textbf{Dimension} \\ \midrule
$\mathcal{D}$         & the set of all diseases                 & 10   \\
$\mathcal{B}$         & the set of all BIO tags for named entities  & 11   \\
$\mathcal{A}$         & the set of all dialogue act categories    & 16   \\
$\mathcal{S}$         & the set of all normalized symptom names      & 331  \\
$\mathcal{L}$         & the set of symptom labels                   & 3    \\
$\mathcal{V}$         & vocabulary of source and target tokens  & 3,138 \\ \botrule
\end{tabular}
\end{table}

Before presenting the experimental results, we first introduce some notations. Let $X = \{x_{1}, x_{2},..., x_{T}\}$ be a piece of doctor-patient dialogue, where $x_i = \{x_{i,1}, x_{i,2},..., x_{i,n}\}$ is the $i$-th utterance, and $x_{i,j} \in \mathcal{V}$ is the $j$-th token in $x_i$. The self-report and the disease category of the patient are denoted as $x_{0}$ and $y \in \mathcal{D}$ respectively, then a medical consultation record can be represented as: $\{x_{0}, X, y\}$. 

The formalization of each task will be introduced in each subsection, and readers can refer to the notations in Table \ref{tab:notation} to better understand our task and evaluation settings. For each task, we only report the baseline models, evaluation metrics and experimental results, the details of experimental settings are provided in the supplementary material.

\subsection{Named Entity Recognition}

\paragraph{Task formalization} ~ {Robust medical named entity recognition (NER) is the first step in understanding doctor-patient conversations. The NER task is designed to automatically predict the boundaries and categories of predefined medical named entities contained in the dialogue. Formally, the NER task aims to predict the BIO label $b_{i}^{j} \in \mathcal{B}$ for each token $x_{i,j}$ given the utterance $x_i$. }

\paragraph{Experimental settings} ~ {We use several popular Chinese named entity models as baselines, including: 1) Lattice LSTM \citep{zhang-yang-2018-chinese}, an extension of Char-LSTM that incorporates lexical information into native LSTM; 2) BERT \citep{devlin-etal-2019-bert}, a bidirectional Transformer encoder with large-scale language pre-training; 3) ERNIE \citep{zhang-etal-2019-ernie}; an improved BERT that adopts entity-level masking and phrase-level masking during pre-training; 4) FLAT \citep{li-etal-2020-flat}, a flat-lattice Transformer that converts the lattice structure into a flat structure consisting of spans; 5) LEBERT \citep{liu-etal-2021-lexicon}, a lexicon enhanced BERT for Chinese sequence labelling, which integrates external lexicon knowledge into BERT layers by a lexicon adapter layer; 6) MC-BERT \citep{zhang-etal-2022-cblue}, a BERT model pre-trained on 20M chinese biomedical sentences using whole entity masking and whole span masking; 7) ERNIE-Health \citep{zhang-etal-2019-ernie}, a language model based on ERNIE pre-trained on 126.9G biomedical chinese dataset, including medical dialogues, scientific articles on medicine and healthcare, electronic medical records (EMRs) and electronic textbooks on medicine and clinical pathology. 

For evaluation, we report token-level metrics, including Precision (P), Recall (R) and F1 Score (F1).}

\paragraph{Experimental Results} ~ {In Table \ref{tab:ner}, we present the experimental results of the NER task. All baselines achieve F1 scores around 90\%. Among them, ERNIE-Health has the highest Precision and F1 score, while LEBERT performs best on Recall. The domain specific language models (LMs) demonstrate a weak advantage over the generic domain LMs. The decent performance of the metrics show that the NER task for medical dialogue is highly feasible in our settings.}

\begin{table}
\centering
\small
\caption{Experimental results for medical NER task. The up arrows and down arrows indicate that the higher the better and the lower the better for the number in the column respectively. All the numbers are percentage values, with the highest value highlighted. It is the same for other tables of experimental results. }
\label{tab:ner}
\begin{tabular}{lccccccccc} \toprule
\textbf{Models}                             & \textbf{P} $\uparrow$     & \textbf{R} $\uparrow$     & \textbf{F1} $\uparrow$    \\ \midrule
Lattice LSTM \citep{zhang-yang-2018-chinese}       & 89.37            & 90.84          & 90.10         \\
BERT-CRF \citep{devlin-etal-2019-bert}             & 88.46                     & 92.35          & 90.37          \\
ERNIE \citep{zhang-etal-2019-ernie}                & 88.87                     & 92.27          & 90.53 \\
FLAT \citep{li-etal-2020-flat}                     & 88.76                     & 92.07          & 90.38          \\
LEBERT \citep{liu-etal-2021-lexicon}               & 86.53                     & \textbf{92.91} & 89.60       \\
MC-BERT \citep{zhang-etal-2022-cblue}  & 88.92                     & 92.18   & 90.52      \\ 
ERNIE-Health \citep{zhang-etal-2019-ernie}    & \textbf{89.71}  &  2.82   &  \textbf{91.24}     \\ \botrule
\end{tabular}
\end{table}

\subsection{Dialogue Act Classification}

\paragraph{Task formalization} ~ {Dialogue act (DA) directly reflects the fine-grained intention of the speaker. Formally, the goal of DAC task is to identify the DA category of each utterance, i.e., to predict the DA label $a_{i} \in \mathcal{A}$ for each utterance $x_i$.}

\paragraph{Experimental settings} ~ {DAC is a typical text classification task. Baseline models include non-pre-trained models: TextCNN \citep{chen2015convolutional}, TextRNN \citep{10.5555/3060832.3061023}, TextRCNN \citep{lai2015recurrent} and DPCNN \citep{johnson2017deep}, generic domain pre-trained models: BERT \citep{devlin-etal-2019-bert} and ERNIE \citep{zhang-etal-2019-ernie} and biomedical pre-trained models: MC-BERT \citep{zhang-etal-2022-cblue} and ERNIE-Health \citep{zhang-etal-2019-ernie}. 

We report 4 metrics for evaluations, including Precision (P), Recall (R), F1 Score (F1) and Accuracy (Acc).}

\paragraph{Experimental Results} ~ {From the results in Table \ref{tab:dac}, it can be seen that the pre-trained model has obvious advantages over the traditional neural models in DAC task, and the benefits brought by the domain specific models are slightly weak. The ERNIE-Health model achieves the best results in the classification task, with the classification accuracy of 82.37\% achieved. The performance of MC-BERT is not as expected both in NER and DAC task, which may be related to the size and distribution of pre-training data.}

\begin{table}
\centering
\small
\caption{Experimental results for DAC task.}
\label{tab:dac}
\begin{tabular}{lcccc} \toprule
\textbf{Models}             & \textbf{P} $\uparrow$  & \textbf{R} $\uparrow$  & \textbf{F1} $\uparrow$  & \textbf{Acc} $\uparrow$  \\ \midrule
TextCNN \citep{chen2015convolutional}   & 74.02              & 70.92           & 72.22             & 78.99             \\
TextRNN \citep{10.5555/3060832.3061023}   & 73.07              & 69.88           & 70.96             & 78.53             \\
TextRCNN \citep{lai2015recurrent}  & 73.82              & 72.53           & 72.89             & 79.40              \\
DPCNN \citep{johnson2017deep}     & 74.30               & 69.45           & 71.28             & 78.75             \\
BERT \citep{devlin-etal-2019-bert}      & 75.35              & 77.16           & 76.14             & 81.62             \\
ERNIE \citep{zhang-etal-2019-ernie}     & \textbf{76.18}     & 77.33  & 76.67  & 82.19    \\ 
MC-BERT \citep{zhang-etal-2022-cblue}     & 75.03     & 77.09  & 75.94    & 81.54    \\ 
ERNIE-Health \citep{zhang-etal-2019-ernie}     & 75.81     & \textbf{77.85}  & \textbf{76.71}    & \textbf{82.37}    \\ 
\botrule
\end{tabular}
\end{table}

\begin{table}
\centering
\small
\caption{Experimental results of symptom recognition in SLI-EXP and SLI-IMP task. The value of hamming loss (HL) is multiplied by $1e4$.}
\label{tab:SLI-SR}
\begin{tabular}{clcccccc}  \toprule
\multirow{2}{*}{\textbf{Task}} & \multirow{2}{*}{\textbf{Models}} & \multicolumn{3}{c}{\textbf{Example Level}} & \multicolumn{3}{c}{\textbf{Label Level}} \\
                             &     & \textbf{SA} $\uparrow$   & \textbf{HL} $\downarrow$  & \textbf{HS} $\uparrow$  & \textbf{P} $\uparrow$        & \textbf{R} $\uparrow$       & \textbf{F1} $\uparrow$       \\ \midrule
\multirow{2}{*}{SLI-EXP}   &  BERT-MLC                          & 73.24             & 10.10          & 84.58          & 86.33            & \textbf{93.14}        & 89.60           \\
                            &   MC-BERT-MLC                     & \textbf{75.34}    & \textbf{9.31}  & \textbf{85.10}  & \textbf{88.47}   & 92.72       & \textbf{90.54}           \\
\midrule
\multirow{4}{*}{SLI-IMP}    & BERT-MLC  & 34.16             & 39.52         & 82.22          & 84.98            & 94.81        & 89.63          \\ 
 & MC-BERT-MLC         & 35.14  & 37.84  & 82.78    & 85.41 & \textbf{95.26}        & 90.07          \\ 
 & BERT-MTL                           & 37.24             & 35.32         & 84.49          & 96.05            & 87.04        & \textbf{91.62}          \\ 
 & MC-BERT-MTL                         & \textbf{37.48}             & \textbf{34.98}         & \textbf{85.34}          & \textbf{95.68}            & 87.56        & 91.44          \\ \botrule
\end{tabular}
\end{table}

\begin{table}
\centering
\caption{Experimental results of symptom inference in SLI-IMP task.}
\label{tab:SLI-SI}
\begin{tabular}{clcccc} \toprule
\textbf{Task} & \textbf{Models}     & \textbf{POS}   & \textbf{NEG}                       & \textbf{NS}    & \textbf{F1}    \\ \midrule
\multirow{4}{*}{SLI-IMP} & BERT-MLC    & \textbf{81.25} & 46.53                     & 59.14 & 62.31 \\
                         & MC-BERT-MLC & 80.80 & 41.30 & 58.15 & 60.08 \\
                         & BERT-MTL    & 79.64 & \textbf{53.87}                     & \textbf{60.20} & \textbf{64.57} \\
                         & MC-BERT-MTL & 80.42 & 53.15 & 59.74 & 64.27 \\ \botrule
\end{tabular}
\end{table}

\subsection{Symptom Label Inference}

Symptom features are the key information to describe the patient's health condition and also the structured training data required for DDP task. The goal of the SLI task is to identify the patient's symptom features from the self-report and the dialogue, and it consists of two cognate sub-tasks: SLI-EXP and SLI-IMP.

\paragraph{Task formalization} ~ {The SLI-EXP task aims to find out the patient's self-provided symptoms in self-report $x_{0}$, which is called \emph{explicit symptoms}, denoted by $\{s_{1}, .. ., s_{k}\}$, where $s_{i} \in \mathcal{S}$. In the contrast, the SLI-IMP task aims to find the symptoms and the corresponding labels in the dialogue $X$, which is called \emph{implicit symptoms}, denoted by $\{(s_{k+1},l_{k+1}), ..., (s_{n},l_{n})\}$, where $s_{j} \in \mathcal{S}$ and $l_{j} \in \mathcal{L}$. Compared with the SLI-EXP task, the SLI-IMP task not only needs to identify symptoms, but also predict the labels of symptoms. We do not need to predict symptom labels in the SLI-EXP task because symptoms in self-report are always positive.}

\paragraph{Experimental settings} ~ {We treat the SLI task as a multi-label classification (MLC) problem, where the label space is $\mathcal{S}$ for SLI-EXP task and $\mathcal{S} \times \mathcal{L}$ for SLI-IMP task. We use BERT \citep{devlin-etal-2019-bert} and MC-BERT \citep{zhang-etal-2022-cblue} as the encoder and obtain the latent vector of the self-report or the entire conversation, which is then mapped into the label space using an MLP layer. The training objectives are the binary Cross-Entropy loss between sigmoid activations of MLP outputs and actual labels. The model is denoted as BERT-MLC and MC-BERT-MLC. 

For SLI-IMP task, we additionally propose a multi-task learning (MTL) \citep{zhang2018overview} based model called BERT-MTL (or MC-BERT-MTL) that can utilize the BIO labels in NER during training. BERT-MTL has three additional MLP layers on top of BERT \citep{devlin-etal-2019-bert}. The role of these three MLP layers is to predict the BIO label of each token, the normalized name of each symptom entity, and the label of each symptom entity. For the first MLP, the input is the hidden vector of each token obtained by BERT encoder, and for the latter two MLPs, the input is the average hidden vector of each symptom entity. The output label space of the three MLP layers are $\mathcal{B}$, $\mathcal{S}$ and $\mathcal{L}$, respectively. The three objectives are trained simultaneously to push the hidden vector of symptom entities to contain more contextual information. We provide the structure diagram of BERT-MTL model in the supplementary material.

We evaluate symptom recognition and symptom inference separately. For the evaluation of symptom recognition, we only focus on whether the mentioned symptom entities are found and the symptom labels are ignored. We report two categories of metrics for multi-label classification, including subset accuracy (SA), hamming loss (HL) and hamming score (HS) in example-based metrics, and precision (P), recall (R) and F1-score (F1) in label-based metrics \citep{zhang2013review}. For symptom inference, we report the macro F1 score (F1) only for those entities that are correctly predicted, F1 scores for each symptom label (\emph{POS, NEG, NS}) are also reported.}

\paragraph{Experimental Results} ~ {The performance of the SLI task is listed in Table \ref{tab:SLI-SR} and \ref{tab:SLI-SR}. For symptom recognition, the performance of subset accuracy (SA) shows that the strict prediction of symptoms is very challenging, especially for implicit symptoms from the entire dialogue (only about 37\%). It is probably due to the exponential growth of the prediction space, as up to dozens of symptoms can be mentioned in a single dialogue. In non-strict cases, both SLI-EXP and SLI-IMP tasks can achieve good performance, with label-level F1 scores reach about 90\%. Moreover, the BERT-MTL model that utilizes BIO labels obtains slight better performance in the symptom recognition in SLI-IMP task, which is also intuitive. Compared with BERT, MC-BERT has some weak advantages in symptom recognition.

For symptom inference in SLI-IMP task, MTL based models have a obvious advantage in the identification for the \emph{NEG} and \emph{NS} categories of symptoms, while MC-BERT seems to have no positive effect. It can be seen that inferring the two categories of symptoms is obviously harder than \emph{POS}, since it often requires more contextual information. Especially for the symptoms that appear in the doctor's utterances, it is likely that the patient's response needs to be observed and analyzed to determine the labels. The results suggest that more efforts are needed to improve the symptom inference.}

\subsection{Medical Report Generation}

\begin{table}
\small
\centering
\caption{Experimental results for MRG task.}
\label{tab:mrg}
\begin{tabular}{lccccccc} \toprule
\textbf{Models} & \textbf{R-1} $\uparrow$  & \textbf{R-2} $\uparrow$   & \textbf{R-L} $\uparrow$   & \textbf{C-F1} $\uparrow$  & \textbf{RD-Acc} $\uparrow$ \\ \midrule
Seq2seq \citep{nallapati-etal-2016-abstractive}      & 54.15         & 38.86          & 50.89          & 35.46          & 39.33          \\
PG \citep{see2017get}          & 57.27         & 43.41          & 53.64          & 43.51          & 53.51  \\
Transformer \citep{vaswani2017attention}     & 53.99         & 39.38          & 49.78          & 37.19          & 45.75           \\
T5 \citep{xue-etal-2021-mt5}          & 60.97 & 44.18 & 57.63 & 47.35          & 49.32           \\ 
ProphetNet \citep{qi2021prophetnet}   & 60.48         & 45.73  & 56.41  & 49.48 & \textbf{61.90}   \\
Bio-ProphetNet   & \textbf{61.83}         & \textbf{47.12}  & \textbf{57.48}          & \textbf{50.12} & 61.15 \\ 
\botrule
\end{tabular}
\end{table}

\paragraph{Task formalization} ~ {The medical report is the summarized patient profile according to the medical consultation record. Formally, the MGR task aims to generate a piece of text $R = \{r_{1}, ..., r_{m}\}$ based on the self-report and the dialogue, where $r_{i} \in \mathcal{V}$.}

\paragraph{Experimental settings} ~ {We treat the MGR task as a text-to-text generation problem. The baseline models include: 1) Seq2seq \citep{nallapati-etal-2016-abstractive}, a LSTM-based encoder-decoder model with attention mechanism; 2) Pointer-Generator (PG) \citep{see2017get}, a improved Seq2Seq model that allows tokens from the source to be directly copied during decoding; 3) Transformer \citep{vaswani2017attention}, the basic model most commonly used in pre-training that based solely on attention mechanisms; 4) T5 \citep{xue-etal-2021-mt5}, a unified Text-to-Text Transformer pre-trained on large text corpus; 5) ProphetNet \citep{qi2021prophetnet}, a large-scale pre-trained generative Transformer based on future prediction strategies; 6) Bio-ProphetNet, a biomedical generative language model based on ProphetNet that we pretrain on the MedDialog dataset. 

 
We measure model performance on standard metrics of ROUGE scores \citep{lin2004looking} that widely used for evaluating automatic summarization task, including ROUGE-1/2/L (R-1/2/L). Besides, we also report Concept F1 score (C-F1) \citep{joshi-etal-2020-dr} to measure the model’s effectiveness in capturing the medical concepts that are of importance, and Regex-based Diagnostic Accuracy (RD-Acc), to measure the model’s ability to judge the disease. To compute C-F1, we use the medical entity extractor (BERT-CRF) trained in our NER task to match entities in the predicted summary to the gold summary, where medical entities in the predicted summary that are not present in the original medical report would be false positives and vice versa for false negatives. For RD-Acc, we use the same regex-based approach mentioned in Section \ref{section:mra}.}

\paragraph{Experimental results} ~ {The results in Table \ref{tab:mrg} illustrate that pre-trained generative models can improve the ROUGE scores of medical reports, but the improved R-2 score compared to PG is quite limited. The improvement of the C-F1 score implies a stronger ability of the the pre-trained model to capture medical concepts in the dialogue. Despite a high score on Rouge, T5 performs mediocrely on D-Acc. Overall, ProphetNet \citep{qi2021prophetnet} has the best performance, which may benefit from the pre-training on large-scale Chinese corpus and the future prediction strategies during decoding. Although pre-trained models improve the fluency of the generated texts, there are still great challenges in scenarios that are highly dependent on knowledge and reasoning.}


\subsection{Diagnosis-oriented Dialogue Policy}

\paragraph{Task formalization} ~ {Different from the above tasks, the DDP task is dynamic task that requires interaction with the \emph{patient simulator} $\mathcal{P}$. Given the patient's explicit symptoms, the goal of the DDP task is to collect the patient's implicit symptoms and predict the disease, within a given maximum number of interactions with $\mathcal{P}$. 

In this article, the patient simulator $\mathcal{P}$ follows the design of \citep{wei2018task}. It can be treated as a function, given the patient's id, and any symptom $s \in \mathcal{S}$ as input, $\mathcal{P}$ can output the patient's symptom label. An \emph{Unknown (UNK)} label will be returned if the symptom $s$ does not appear in the dialogue.

More specifically, the agent asks $\mathcal{P}$ for one symptom at each step, and after receiving a feedback, asks the next symptom, and repeating above for several turns, until the agent obtains enough information to make diagnosis.

The patient simulator can be designed to be more practical, i.e., the input and output are both natural language texts. In this case, we need a language transmitter to generate the text that conveys the semantics of the action selected by the agent, and a language interpreter based on the proposed understanding framework to parse the patient's response. This situation is one of our future research directions, but it is beyond the scope of this article since our DDP task focuses on policy learning with structured data.

}

\paragraph{Experimental settings} ~ {Baseline models include DQN \citep{wei2018task}, KR-DQN \citep{xu2019end}, REFUEL \citep{kao2018context}, GAMP \citep{xia2020generative} and HRL \citep{zhong2022hierarchical}. Except for GAMP, all other methods are based on reinforcement learning (RL). In RL settings, at each turn of interaction, the agent chooses an action from the joint action space of all symptoms and diseases, and correct symptom queries and disease diagnoses are positively rewarded, then the policy can be learned by maximizing the empirical expected cumulative reward. In the contrast, GAMP is a GAN-based method that uses the GAN network to avoid generating randomized trials of symptom, and adds mutual information to encourage the model to select the most discriminative symptoms. We set the maximum number of interactions between all agents and the patient simulator to 10. 




Ideally, if the agent collects all the implicit symptoms, the disease classifier can utilize all the symptom features of the patient. In this case, the performance of disease classification is intuitively the best. We train support vector machine (SVM) \citep{noble2006support} classifier with the complete symptoms (both explicit and implicit symptoms), and call the model upper-bound-SVM (UB-SVM). It is an invalid static agent, but its performance can provide a certain reference for dynamic agents.


To evaluate the agent, we report three most concerned metrics, namely symptom recall (SX-Rec), diagnostic accuracy (DX-Acc), and average number of turns (\# Turns). Symptom recall measures the agent's ability to find implicit symptoms of the patient, diagnostic accuracy measures the agent's ability in disease classification, and average number of turns indicates the efficiency of the diagnostic process.}

\paragraph{Experimental results} ~ {From the results in Table \ref{tab:ddp}, HRL obtains the best symptom recall and diagnostic accuracy compared to other baselines, with an acceptable average number of turns. HRL groups diseases and works in a combination of master and multiple workers, which is more in line with the actual medical division of labor. However, the symptom recall and diagnostic accuracy of existing models are still far from acceptable levels. It is worth noting that the SVM-UB model can achieve a diagnostic accuracy of 70\%, suggesting that the performance of dynamic agents can be expected to be improved if the agents is able to find more implicit symptoms.}




\begin{table}
\centering
\small
\caption{Experimental results for DDP task.}
\label{tab:ddp}
\begin{tabular}{lccc} \toprule
\textbf{Models} & \textbf{SX-Rec} $\uparrow$ & \textbf{DX-Acc} $\uparrow$ & \textbf{Avg. \# Turns} \\ \midrule
UB-SVM \citep{noble2006support}   & -  & 0.706 & -     \\ \midrule
DQN \citep{wei2018task}             & 0.047           & 0.408         & 9.75             \\
KR-DQN \citep{xu2019end}         & 0.279           & 0.485          & 6.75             \\
REFUEL \citep{kao2018context}          & 0.262           & 0.505         & 5.50             \\
GAMP \citep{xia2020generative}            & 0.067           & 0.500         & 1.78            \\
HRL \citep{zhong2022hierarchical}            & \textbf{0.295}  & \textbf{0.556} & 6.99     \\  \botrule
\end{tabular}
\end{table}

\section{Conclusions}
\label{section:conclusion}

In this article, we propose a design of frameworks and tasks for automatic medical consultation system to support both static and dynamic medical scenarios. We introduce a new medical dialogue dataset called IMCS-21 with multi-level fine-grained annotations and establish five tasks under the proposed framework. We develop widely used neural-based models for each task and demonstrate experimental results to give an insight about the performance of different tasks. The experimental results show that the validity and potential of the corpus make it expected to be an important benchmark for automated medical consultation systems.


\section{Funding}

This work is partially supported by Natural Science Foundation of China (No.71991471, No.6217020551), Science and Technology Commission of Shanghai Municipality Grant (No.20dz1200600, 21QA1400600) and Zhejiang Lab (No. 2019KD0AD01).

\emph{Conflict of Interest:} The authors declare no competing interests.

\bibliographystyle{natbib}
\bibliography{document}

\end{document}